\documentclass{article}

    \PassOptionsToPackage{numbers, compress}{natbib}

     \usepackage[final]{neurips_2019}

\usepackage[utf8]{inputenc} 
\usepackage[T1]{fontenc}    
\usepackage{hyperref}       
\usepackage{url}            
\usepackage{booktabs}       
\usepackage{amsfonts}       
\usepackage{nicefrac}       
\usepackage{microtype}      

\usepackage{wrapfig}

\usepackage{amssymb}
\usepackage{amsmath}
\usepackage{graphicx}
\usepackage{subcaption}
\usepackage{cleveref}

\usepackage{bm}
\newcommand{\vect}[1]{\bm{#1}}
\newcommand{\matr}[1]{\bm{#1}}

\newcommand{\vh}[0]{\vect{h}}
\newcommand{\vv}[0]{\vect{v}}
\newcommand{\vx}[0]{\vect{x}}

\newcommand{\vp}[0]{\vect{p}}

\newcommand{\mW}[0]{\matr{W}}

\usepackage{url}

\title{Can adversarial training learn image captioning ?}

\author{
    Jean-Benoit Delbrouck
  \And
  Bastien Vanderplaetse
  \And
   St\'ephane Dupont
  \and
  ISIA Lab, Polytechnic Mons, Belgium \\
  \and
  \{jean-benoit.delbrouck, bastien.vanderplaetse, stephane.dupont\}@umons.ac.be
  \\
}

\begin{document}

\maketitle

\begin{abstract}
Recently, generative adversarial networks (\textsc{GAN}) have gathered a lot of interest. Their efficiency in generating unseen samples of high quality, especially images, has improved over the years. In the field of Natural Language Generation (\textsc{NLG}), the use of the adversarial setting to generate meaningful sentences has shown to be difficult for two reasons: the lack of existing architectures to produce realistic sentences and the lack of evaluation tools. In this paper, we propose an adversarial architecture related to the conditional \textsc{GAN} (\textsc{cGAN}) that generates sentences according to a given image (also called image captioning). This attempt is the first that uses no pre-training or reinforcement methods. We also explain why our experiment settings can be safely evaluated and interpreted for further works.
\end{abstract}

\section{Introduction}
Generative adversarial networks (GAN, \cite{goodfellow2014generative}) have attracted a lot of attention over the last years especially in the field of image generation. GAN have shown great success to generate high fidelity, diverse images with models learned directly from data. Recently, new architectures have been investigated to create class-conditioned GAN \cite{brock2018large} so that the model is able to generate a new image sample from a given ImageNet category. These networks are more broadly know as conditional-GAN or cGAN \cite{mirza2014conditional} where the generation is conditioned by a label. \\

In the field of Natural Language Generation (NLG), on the other hand, a lot of efforts have been made to generate structured sequences. In the current state-of-the-art, Recurrent neural networks (RNN; \cite{graves2013generating}) are trained to produce a sequence of words by maximizing the
likelihood of each token in the sequence given the current (recurrent) state and the previous token. Scheduled sampling \cite{bengio2015scheduled} and reinforcement learning \cite{zoph} have also been investigated to train such networks. Unfortunately, training discrete probabilistic models with GAN has shown to be a very difficult task. Previous investigations require complicated training techniques such as gradient policy methods and pre-training and often struggles to generate realistic sentences. Moreover, it is not always clear how NLG should be evaluated in an adversarial settings \cite{semeniuta2018accurate}. \\

In this paper, we propose a cGAN-like architecture that generates a sentence according to a label, the label being an image to describe. This work is related to image captioning task that proposes strict evaluation methods for any given captioning data-set. We also investigate if GAN can learn image captioning in a straightforward manner, this includes a fully differentiable end-to-end architecture and no pre-training. The generated sentences are then evaluated against to the ground truth captioning given by the task. The widely-used COCO caption data-set  \cite{mscoco} contains 5 human-annotated ground-truth descriptions per image, this justifies our will to use a generative adversarial setting whose goal is to generate realistic and diverse samples.

\section{Related work}

A few works can be related to ours. First, \cite{yu2017seqgan} proposed a Sequence Generative Adversarial Nets trained with policy gradient methods \cite{sutton2000policy} and used synthetic data experiments to evaluate the training. Other works also investigated adversarial text generation with reinforcement learning and pretraining \cite{guo2018long, dai2017towards}. Finally, the closest work related to ours is the one of \cite{press2017language} who proposes an adversarial setting pre-training and without reinforcement. Our model differs in the way that we use a conditional label as image to generate a sentence or image caption. 

\section{Adversarial image captioning}
In this section, we briefly describe the model architecture used in our experiments. 

As any adversarial generative setting, our model is composed of a generator $G$ and a discriminator $D$. The generator $G$ is an RNN that uses a visual attention mechanism \cite{pmlr-v37-xuc15} over an image $I$ to generate a distribution of probabilities $p_t$ over the vocabulary at each time-step $t$. During training, $G$ is fed a caption as the embedded ground-truth words and $D$ is fed with either the set of probability distributions from $G$ or the embedded ground truth words of a real caption. $D$ has to say if the input received is either real or fake according to the image. $D$ is also a the same RNN as $G$ but with different training weights. The RNN can be expressed as follows:

\begin{align} \label{eq:pt}
\text{if $G$}: \vh_t =& ~ f_{\text{gru}_1}(\vx_t,\vh^\prime_{t-1}) \nonumber \\
\text{if $D$}: \vh_t =& ~ f_{\text{gru}_1}(\vx_t \, \text{or} \, \vp_t,\vh_{t-1}) 
\end{align}
\begin{align}
    \vv_t =& ~f_{\text{att}}(\vh_t,I)  \\
     \vh_t^\prime =& ~ f_{\text{gru}_2}(\vh_t,\vv_{t})
\end{align}
\begin{align}
\text{if $G$}: \vp_t =& ~ \text{softmax}(\mW^{\text{proj}} \vh_t^\prime)  \nonumber \\
\text{if $D$}: [0,1] =& ~ \mW^{\text{ans}} \vh_t^\prime
\end{align}

where $x_t$ is the embedded ground-truth symbols of word $t$ and $f_{\text{att}}$ the attention model over image $I$.
$G$ and $D$ are both trained simultaneously with the following min-max objective:

\begin{align}
    \min_{G} \max_{D} \mathbb{E}_{x \sim \mathbb{P}_r}[ & \log (D(x))] \nonumber \\
    & +  \mathbb{E}_{\tilde{x} \sim \mathbb{P}_g}[\log (1 - D(\tilde{x}))] 
\end{align}

where $x$ is an example from the true data and $\tilde{x} = G(z)$ a sample from the Generator. Variable $z$ is supposed to be Gaussian noise. 

\section{Tips and tricks}

It is important to mind two tricks to make adversarial captioning work:

\textbf{Gradient penality for embeddings} \quad As show in equation \ref{eq:pt}, the discriminator receives half of the time a probability distribution over the vocabulary from G. This is fully differentiable compared to $\text{arg\,max} \,p_t$. A potential concern regarding our strategy to train our discriminator
to distinguish between sequence of 1-hot vectors from the true data distribution and a sequence of probabilities from the generator is that the discriminator can easily exploit the sparsity in the 1-hot vectors. However,  a gradient penalty can be added to the discriminator loss to provides good gradients even under an optimal discriminator. The gradient penalty \cite{wgangp.article} is defined as $ \lambda \mathbb{E}_{ \hat x \sim \mathbb{P}_{\hat x}}[(\Vert {\nabla_{\hat x}D(\hat x)}\Vert_2-1)^2]$ with $\hat x = \epsilon x + (1-\epsilon) \tilde{x}$ and where $\epsilon$ is a random number sampled from the uniform distribution $U[0,1]$
    
\textbf{Dropout as noise} \quad For the evaluation of a model to be consistent, we can't introduce noise as input of our Generator. To palliate this constraint, we provide noise
only in the form of dropout to make our Generator less deterministic. Because we don't want to sample from a latent space (our model don't fall into the category of generative model), using only dropout is a good work-around in our case. Moreover, dropout has already shown success in previous generative adversarial work \cite{isola2017image}.

\section{Experimentation} \label{sec:exp}
We use the MS-COCO data-set \cite{mscoco}consisting of 414.113 image-description pairs. For our experiments, we only pick a subset of 50.000 training images, 1000 images are use for validation.

Each ground-truth symbol $\vx_t \in \mathbb{R}^{300}$ is a word-embedding from Glove \cite{glove.article}. All GRU used are of size 256, so is $\vh_t$. Image $I$ is extracted at the output of the \textit{pool-5} layer from ResNet-50 \cite{resnet.article}. The attention mechanism $f_{\text{att}}$ consists of a simple element-wise product between $vh_t$ and $I$ :
$$\vv_t = \vh_t \odot \mW_I $$
where $\mW_I \in \mathbb{R}^{2048\times256}$  and $\vv_t \in \mathbb{R}^{256}$. Finally, the size of the following matrices are: $\mW^{\text{proj}} \in \mathbb{R}^{256\times|\mathbb{V}|}$ where $|\mathbb{V}|$ is the vocabulary size and $\mW^{\text{ans}} \in \mathbb{R}^{256\times1}$.

As hyper-parameters, we set the batch size to 512, the gradient penalty $\lambda = 9$ and a dropout of p=0.5 is applied at the output of $f_{\text{gru}_1}$ in the Generator. We stop training of the BLEU score on the validation set doesn't improve for 5 epochs.

\section{Results}
\begin{figure} [h]
    \centering
		\begin{subfigure}[b]{0.3\textwidth}
			\centering \includegraphics[width=\textwidth]{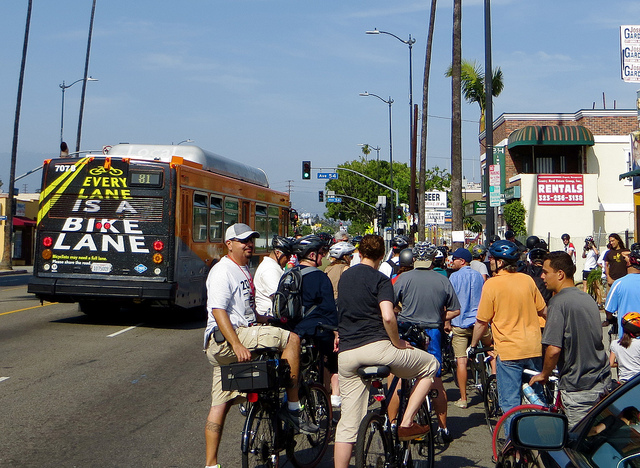}
			\caption{Ground truth : \textit{a group of people who are sitting on bikes} \\ Generated caption : \textit{a group of people riding on the side of a car} \\ \textsc{BLEU}$-4$ $=$ $0.683$}
		\end{subfigure}
		\begin{subfigure}[b]{0.3\textwidth}
			\centering \includegraphics[width=\textwidth]{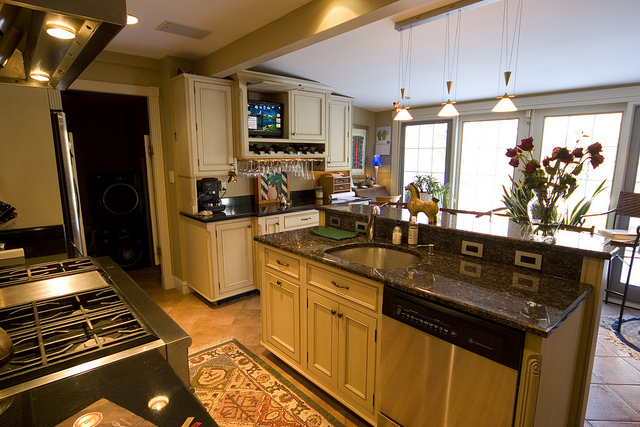}
			\caption{Ground truth : \textit{a kitchen with a stove a sink and a counter} \\ Generated caption : \textit{a kitchen with a sink stove a sink and other} \\ \textsc{BLEU}$-4$ $=$ $0.719$}
		\end{subfigure}
		\begin{subfigure}[b]{0.3\textwidth}
			\centering \includegraphics[width=\textwidth]{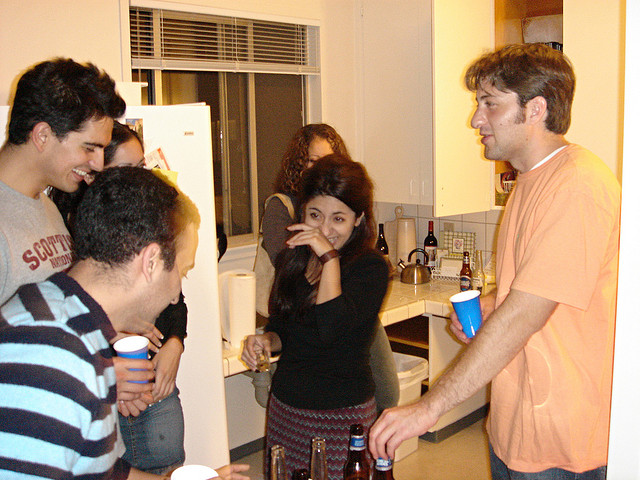}
			\caption{Ground truth : \textit{a group of people standing around a kitchen} \\ Generated caption : \textit{a group of people standing in a kitchen} \\ \textsc{BLEU}$-4$ $=$ $0.946$}
		\end{subfigure}
		\caption{Success case of our adversarial captioning model. The model is able to recognize groups of people, some locations and objects. We also notice the correct use of verbs.}
		\label{fig:best}
\end{figure}

The best configuration as described in section \ref{sec:exp} gives a \textsc{BLEU}$-4$ score \cite{bleu.article} of 7.30.  Figure \ref{fig:best} shows some of the best generated captions given images. We observed that the model is able to recognize groups of people as well as some locations (such as a  kitchen) and objects (such as a sink). The model also learned to use the correct verb for a given caption. For example, in Figure \ref{fig:best} the model is capable of making differentiate \textit{riding} with \textit{standing}. 

\begin{figure} [h]
    \centering
		\begin{subfigure}[b]{0.3\textwidth}
			\centering \includegraphics[scale=0.15]{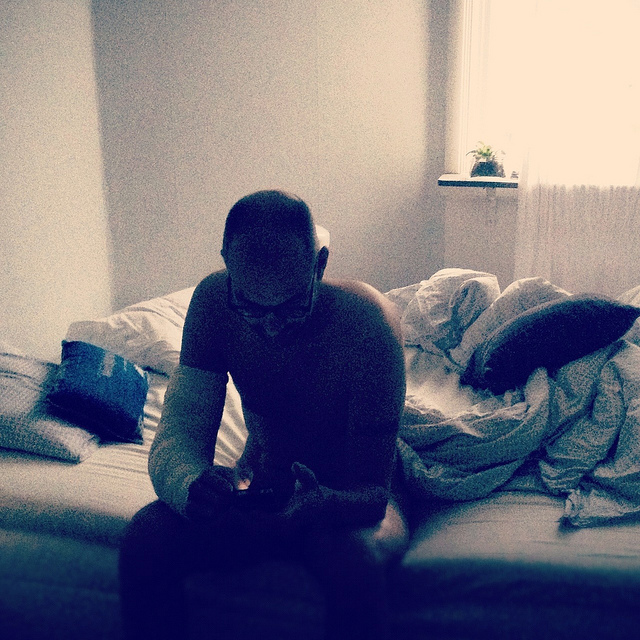}
			\caption{Ground truth : \textit{a nude man sitting on his bed while using his phone} \\ Generated caption : \textit{a <unk> <unk> <unk> next to an table}}
		\end{subfigure}
		\begin{subfigure}[b]{0.3\textwidth}
			\centering \includegraphics[width=\textwidth]{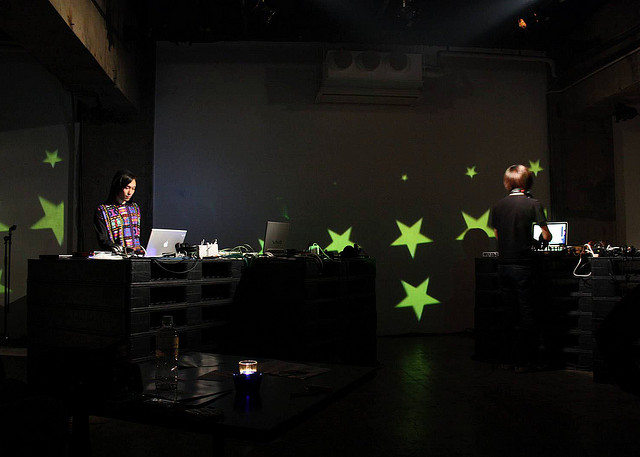}
			\caption{Ground truth : \textit{two people stand using laptops in a dark room with big stars on the wall} \\ Generated caption : \textit{a kitchen with a sink and tiled sink}}
		\end{subfigure}
		\begin{subfigure}[b]{0.3\textwidth}
			\centering \includegraphics[width=\textwidth]{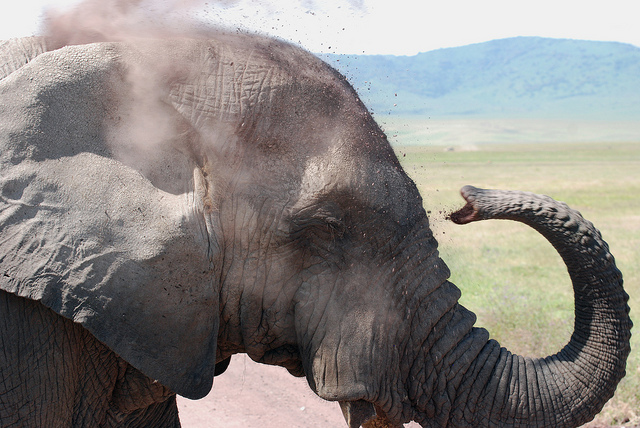}
			\caption{Ground truth : \textit{an elephant using its trunk to blow the dirt off its face} \\ Generated caption : \textit{a man of people sits in a kitchen}}
		\end{subfigure}
		\caption{Worst generated captions (\textsc{BLEU}$-4$ $=$ $0$)}
		\label{fig:worst}
\end{figure}

Nevertheless, we can identify two failure cases. First, the model often output sentences filled with the <unk> token. It is possible that the model hasn't been trained for long enough and on too few data. The Generator receives only a single adversarial feed back for all the words generated. It is possible some words may not have received enough gradient in order to be successfully used. In general, the pool of words used is not very large: the words used in Figure \ref{fig:best} are related to the ones used in Figure \ref{fig:worst}. Secondly, the model sometimes outputs well formed sentences (Figure \ref{fig:worst} b) and c)) but unrelated to the image. Here, it is possible that the conditional information has not been taken into account.

\section{Conclusion}

\begin{wrapfigure}{l}{0.5\textwidth} 
    \includegraphics[scale=0.4]{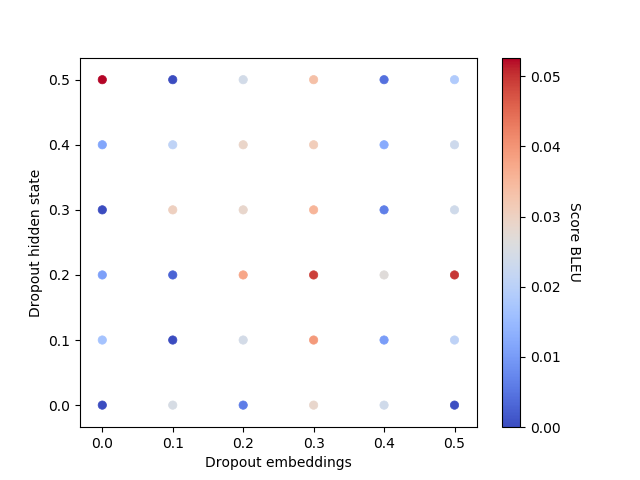}
    \caption{Result of the dropout on embedding and hidden state}
    \label{fig:do}
\end{wrapfigure}

In this paper, we made a first attempt on adversarial captioning without pre-training and reinforcement techniques. The task is challenging, especially since the generator G and discriminator D work with different sparsity. Nevertheless, only the WGAN with gradient penalty was able to give acceptable results. Other techniques such as the relativistic GAN \cite{jolicoeur2018relativistic} or WGAN-divergence \cite{wu2018wasserstein} didn't work in our case. We also notice that the model was very sensitive to dropout. However, Figure \ref{fig:do} confirms our intuition that no dropout is not benefical for the generator (the bottom-left of the heat-map resulted in a BLEU score of 0).
\\ \\ \\
There are a few improvements that can be made for future research. First, the attention model could be more sophisticated so that the visual signal is stronger. The size of the overall model could also be increased. Finally, the model should be trained on the full COCO training set. It is possible that enforcing an early-stop of 5 epochs for training could be an issue since the model could take time to converge.   

\bibliographystyle{unsrt}
\bibliography{nips2019}
\end{document}